\pdfoutput=1
\documentclass[11pt]{article}
\usepackage{color}
\usepackage[dvipsnames]{xcolor}

\usepackage{ACL2023}

\usepackage{times}
\usepackage{soul}
\usepackage{latexsym}
\usepackage{amsthm}
\usepackage{amsmath}
\usepackage{amsfonts}
\usepackage{mathtools}
\usepackage{amssymb}
\usepackage{booktabs}
\usepackage{multirow}
\usepackage{subcaption}
\usepackage{mwe}
\usepackage{enumitem}
\usepackage[T1]{fontenc}
\usepackage{mdframed}
\usepackage{colortbl}
\usepackage{arydshln}
\usepackage{circledsteps}
\usepackage[utf8]{inputenc}

\usepackage{microtype}

\usepackage{inconsolata}
\usepackage{graphicx}
\usepackage{tikz}
\usetikzlibrary{positioning}
\usetikzlibrary{arrows.meta}

\theoremstyle{definition}
\newtheorem{definition}{Definition}

\theoremstyle{remark}
\newtheorem*{remarks}{Remark}

\definecolor{tabColor}{RGB}{223, 227, 240}

\newcommand{\name}{\textsc{RoRa}}
\title{
\vspace*{-0.5in}
{{\small \hfill ACL'24}\\
\vspace*{.25in}} 
{\name{}:} Robust Free-Text Rationale Evaluation}

\newcommand{\g}{\textit{gold}}
\newcommand{\lky}{\textit{leaky}}
\newcommand{\va}{\textit{vacuous}}
\newcommand{\gl}{\textit{gold}$_\textit{leaky}$}

\author{
Zhengping Jiang$^{\heartsuit}$ \quad 
Yining Lu$^{\heartsuit}$ \quad
Hanjie Chen \\
{\bf 
Daniel Khashabi \quad 
Benjamin Van Durme \quad
Anqi Liu 
}\\
Johns Hopkins University\\
\texttt{\{zjiang31, ylu130\}@jhu.edu}
}

\begin{document}
\maketitle
\def\thefootnote{$\heartsuit$}\makeatletter\def\Hy@Warning#1{}\makeatother\footnotetext{
Equal contribution}
\def\thefootnote{\arabic{footnote}}
\begin{abstract}

% [Ben] Free-text rationales for QA are [define what they are].  REV (cite) was recently proposed as a way to quantify the informativeness of such rationales. We find that REV is highly sensitive to to \emph{leaky information}, [define leaky information].  We propose R2EV as a rationale evaluation solution that removes these biases.

Free-text rationales play a pivotal role in explainable NLP, bridging the knowledge and reasoning gaps behind a model's decision-making. However, due to the diversity of potential reasoning paths and a corresponding lack of definitive ground truth, their evaluation remains a challenge. Existing evaluation metrics rely on the degree to which a rationale \emph{supports} a target label, but we find these fall short in evaluating rationales that inadvertently \emph{leak the labels}. To address this problem, we propose \name{}, a \underline{RO}bust free-text \underline{RA}tionale evaluation against label leakage.\footnote{\url{https://github.com/zipJiang/RORA}} \name{} quantifies the new information supplied by a rationale to justify the label. 
This is achieved by assessing the conditional $\mathcal{V}$-information \citep{hewitt-etal-2021-conditional} with a predictive family robust against leaky features that can be exploited by a small model.
% \daniel{The only place we use "deficient" term and once toward the end of the paper.}
\name{} consistently outperforms existing approaches in evaluating human-written, synthetic, or model-generated rationales, particularly demonstrating robustness against label leakage. We also show that \name{} aligns well with human judgment, providing a more reliable and accurate measurement across diverse free-text rationales.
% \footnote{\href{https://github.com/zipJiang/REV-reimpl}{Code to reproduce our work is available}}
\end{abstract} 

\section{Introduction}

\begin{figure*}
    \centering
    \includegraphics[width=\linewidth]{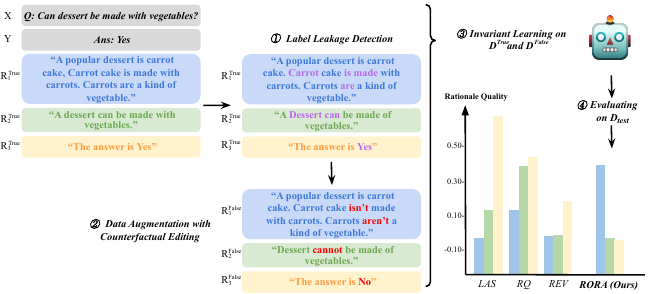}
    \caption{\name{} framework for evaluating rationales $R_1^{\text{True}}$, $R_2^{\text{True}}$, $R_3^{\text{True}}$. Existing baselines are highly sensitive to rationales that \textcolor{orange}{simply restate the label} or \textcolor{Green}{paraphrase the given question and label}, leading to inflated scores compared to the \textcolor{blue}{human-annotated} rationale. In contrast, \name{} provides an informativeness score that better characterizes rationale quality. It is achieved by \Circled{1} detecting potential leakage tokens in the rationale (\S\ref{subsec:leakage-detection}) and \Circled{2} generate additional training data with counterfactual editing for data augmentation (\S\ref{subsec: data augmentation}), followed by \Circled{3} training an evaluation model invariant to label leakage (\S\ref{subsec: evaluation}).}
    \label{figure: pipeline}
\end{figure*}

The ability of large language models (LLMs) to generate free-text rationales that elaborate on their decision-making processes holds promise for explainable NLP, either in the form of a reasoning chain \citep{wei2022chain, yao2023tree} or post-hoc explanations \citep{madaan2023self, zheng2023judging}. Previous works have also collected human-written rationales to enhance model reasoning and the generation of free-text rationales \citep{rajani2019explain, camburu2018esnli, aggarwal-etal-2021-explanations, mor2021strategyqa}.

However, evaluating these rationales remains an open problem because of the diversity of reasoning paths and the lack of definitive ground truth \citep{chan2022frame}. As a result, existing metrics rely on measuring how much the rationale supports a given label. This is usually achieved by comparing predictions of models trained with and without rationales. For example, Leakage-Adjusted Simulatability (LAS) \citep{hase-etal-2020-leakage} and Rationale Quality (RQ) \citep{wiegreffe2022measuring} measure rationale quality through the difference in accuracy. Alternatively, Rationale Evaluation with conditional-$\mathcal{V}$-information (REV) \citep{chen-etal-2023-rev} evaluates the reduction in model predictive uncertainty upon conditioning on the rationale. 

Yet all these methods are vulnerable to \textit{label leakage} \citep{li-etal-2022-systematic, chen2023relation, ludan2023explanationbased}: the rationale inadvertently \textit{paraphrasing} or \textit{restating} labels, creating a spurious shortcut \citep{geirhos2020shortcut} for the evaluation model to infer the label. The critical issue stems from the mismatch in objectives: existing methods evaluate how easy it is to utilize information in the rationale, but rationales are \textit{explanations}, whose quality does not always come with simplicity. The best explanation has to support the answer through some sense of mechanisms, such as methodically considering a set of axioms and running through a deductive chain \citep{bechtel2005explanation, keil2006explanation, glennan2002rethinking}, without which they are mere ``effects'' \citep{cummins2000does}. \autoref{figure: pipeline} shows an example where existing evaluation methods are highly sensitive to label leakages in paraphrased or restated rationales, while in fact, these label leakages merely increase the predictive probability without providing any meaningful explanations.\footnote{Note that scores between different evaluation metrics are not directly comparable because of different scales and criteria. In this paper, our analysis mainly focuses on the ranking and relative differences within each metric.} 

With this objective in mind, we introduce \name{}, a novel approach to evaluate rationales robust to label leakage. \name{}'s construction consists of three stages as illustrated in \autoref{figure: pipeline}. First, we fit a small model and identify label-leaking tokens via its gradient-based attributions (\S\ref{subsec: data augmentation}). After that, we generate additional training data with counterfactual editing \citep{ross-etal-2021-explaining} (\S\ref{subsec: data augmentation}). Finally, we force the evaluation model to ignore these label-leaking tokens through invariant learning \citep{arjovsky2020invariant} (\S\ref{subsec: evaluation}). 
Our approach aligns with the human perception that explanations should apparently increase the understanding of a given phenomenon by helping to create knowledge and to develop better theories \citep{wilson1998shadows}. On the contrary, label leakage tends to be repetitive and tautological \citep{aslanov2023tautological}, dominating the insightful parts of the explanation.

We compare \name{} with baseline metrics \citep{hase-etal-2020-leakage, wiegreffe2022measuring, chen-etal-2023-rev} in evaluating various synthetic and human-annotated rationales, with or without label leakage, on three QA datasets. \name{} consistently outperforms baseline metrics by providing robust evaluations against label leakages. We also compare \name{} against model-generated rationale evaluation and demonstrate its better agreement with human evaluation.

\section{Motivation and Backgrounds}

Following previous work by \citet{chen-etal-2023-rev}, \name{} adopts the framework of conditional $\mathcal{V}$-information \citep{hewitt-etal-2021-conditional}. In this section, We outline the general idea of this framework and point out how the objective of \name{} is critically different from previous attempts.

\subsection{Conditional $\mathcal{V}$-information}
\label{subsec: conditional v-information}
% \yining{mention we also follow v-information}
The theory of $\mathcal{V}$-information, as proposed by \citet{xu2020theory}, focuses on quantifying usable information under computational constraints. Specifically, it examines how much usable information about a random variable $Y$ can be derived from another random variable $X$ by applying functions belonging to a specified set known as the \textit{predictive family} $\mathcal{V}$. Due to its transparent connection to baseline probing, \citet{hewitt-etal-2021-conditional} propose to extend $\mathcal{V}$-information to multivariable cases as an evaluation of information beyond a baseline.

%\begin{definition}[Multivariable Predictive Family]
%Let $\Omega = \{f: \mathcal{X}'_1 \times \dots \times \mathcal{X}'_n \rightarrow \mathcal{P}(\mathcal{Y})\}$, where $\mathcal{X}'_i = \mathcal{X}_i \cup \{\varnothing\}$.\footnote{We adopt the notational conventions of \citet{xu2020theory} (rather than that of \citet{hewitt-etal-2021-conditional}) since it abstracts away actual implementation from the underlying idea.} We say that $\mathcal{V} \subseteq \Omega$ is a predictive family if
%\begin{equation}
%\begin{aligned}
%\forall f, \ x_1, \dots , x_n \in \mathcal{V} \times \mathcal{X}'_1 \times \cdots \times \mathcal{X}'_n, \\
%\exists f' \in \mathcal{V}, s.t. \ \forall c \subset \{x_1,\cdots, x_n\} , \\
%f[x_1, \cdots,x_n] = f'[c, \{\varnothing, \cdots, \varnothing\}].
%\end{aligned}
%\end{equation}
%\end{definition}
%\begin{remarks}
%    The definition ensures that a predictive family can ignore any feature subset while the remaining subset $c$ still has the same output.
%    In our case, it ensures that one can always construct a valid, more limited predictive family by forcing ignorance on leaky features. 
%\end{remarks}

\begin{definition}[Multivariable $\mathcal{V}$-information]
    Let $X_1, X_2 \dots, X_n \in \mathcal{X}_1, \mathcal{X}_2, \dots, \mathcal{X}_n $ and $Y \in \mathcal{Y}$ be random varaibles. Let $\mathcal{V}$ be a multivariable predictive family.\footnote{We refer the readers to the work of \citet{xu2020theory} We adopt their notation (rather than that of \citet{hewitt-etal-2021-conditional}) since it abstracts away actual implementation from the underlying idea.} Then the conditional multivariable $\mathcal{V}$-information from $X_{j}$ to $Y$, where $j \in \{1, 2, \dots, n\}$, conditioned on prior knowledge $C \subset \{X_1, \dots, X_n\}$, is defined as
    \begin{equation}
        I_{\mathcal{V}} (X_l \rightarrow Y | C) = H_{\mathcal{V}}(Y|C) - H_{\mathcal{V}}(Y|C \cup X_l),
    \label{equation: conditional v-information}
    \end{equation}
    where
        \begin{equation}
        % \small
        \begin{aligned}
            H_{\mathcal{V}}(Y|C) &=   \\   \inf_{f\in\mathcal{V}}& \mathbb{E}_{c,y}\Big[ -\log f\big[c, \{\varnothing, \dots, \varnothing\}\big](y)\Big].
        \end{aligned}
        \label{equation: conditional v-entropy}
    \end{equation}
    is the multivariable predictive $\mathcal{V}$-entropy.\footnote{Following \citet{xu2020theory} $\varnothing$ is used to indicate that no information is provided for this feature. In practice, it is common to mask out this feature with constant values \citep{hewitt-etal-2021-conditional}.}
\end{definition}

\begin{remarks}
   \autoref{equation: conditional v-information} is also called the ``conditional $\mathcal{V}$-information'' (and \autoref{equation: conditional v-entropy} ``conditional $\mathcal{V}$-entropy''). \autoref{equation: conditional v-information} characterizes how much uncertainty of a variable $Y$ can be reduced when a certain class of function was used to extract knowledge from observing $X_{l}$. The formulation of \autoref{equation: conditional v-information} allows estimation of mutual information via optimization over $\mathcal{V}$, i.e., through gradient descent. It is important to note that the definition of the predictive family ensures that a predictive family can ignore any feature subset while the remaining subset $c$ still has the same output. In our case, it ensures that one can always construct a valid, more limited predictive family by forcing ignorance on leaky features.
\end{remarks}

\subsection{Evaluating Label-Leaking Rationales}
\label{subsec: evaluating label-leaking rationales}
The conditional $\mathcal{V}$-information provides a natural way to evaluate the informativeness of rationales. For a natural language problem with input $X$ and label $Y$, A direct application of multivariable $\mathcal{V}$-information to evaluate the quality of rationale $R$ is
\begin{equation}
    Q(R) = I_{\mathcal{V}}(R \rightarrow Y | X).
    \label{equation: v-information}
\end{equation}
Yet, as \citet{hase-etal-2020-leakage} pointed out, this approach fails to distinguish between different ways a rationale provides knowledge for a model's prediction. For example, a bad rationale can obtain a very high $Q(R)$ by restating the label to be predicted. To solve this problem, REV \citep{chen-etal-2023-rev} proposes to introduce a vacuous rationale $B$ that is simply a declarative combination of $X$ and $Y$

\begin{equation}
    \text{REV}(R) = I_{\mathcal{V}}(R\rightarrow Y|B).
    \label{equation: rev formula}
\end{equation}
However, this requires that $Y$ is not completely determined by $B$, which can be easily violated by label leakage in $B$ that makes $H_{\mathcal{V}}(Y|B)$ close to 0.\footnote{Conditioning on $B$ may cause other problems, of which we refer the readers to \autoref{Appendix: further discussion on rev and label leakage} for a more detailed discussion.}

Under a similar framework where conditional $\mathcal{V}$-entropy $H_{\mathcal{V}}$ in \autoref{equation: conditional v-information} is replaced by prediction accuracy, other methods are proposed to mitigate the impact of label leakage. \citet{hase-etal-2020-leakage} propose LAS: an averaged accuracy between leaking and non-leaking rationales which are classified by whether a simulator model can predict $Y$ solely from the explanation $R$. However, the prevalence of label leakage in free-text rationales may overwrite the effect of macro-averaging. The metric is simply undefined when all rationales in the dataset leak labels \citep{pruthi2022evaluating}. \citet{sia2022logical} suggest that using logical counterfactuals may circumvent the label leakage issue. However, their approach does not generalize to most NLP tasks because rationales in free-text format are usually too complex to be parsed logically.

Suppose a free-text rationale $R$ consists of label leakage $R_L$ and non-leaky information $R_R$ (i.e., $R = R_L\cup R_R$). We propose that a good evaluator $\mathcal{S}$ of the informativeness of rationale $R$ would be
\begin{equation}
    \begin{aligned}
        \mathcal{S}(R) &= I_\mathcal{V}(R_R \rightarrow Y|X), \\
        &= I_\mathcal{V}(R \setminus R_L \rightarrow Y|X).
    \end{aligned}
    \label{eq:objective}
\end{equation}
Notice that we do not condition on $B$, as in our formulation, vacuous rationales are similar to spurious features about which our evaluator should be ignorant \citep{xu2020theory}. $R_L$ and $R_R$ are usually deeply entangled in natural language utterances, and it is difficult to isolate $R_R$ completely. To address this problem, we propose \name{} to approximate \autoref{eq:objective} by identifying and forcing the model to unlearn $R_L$, as detailed in \S\ref{sec:methods}.

\section{\name{} for Rationale Evaluation}
\label{sec:methods}

We describe our framework (\autoref{figure: pipeline}) to estimate \autoref{eq:objective} as a measurement of the informativeness of rationales.

\subsection{Leakage Detection}
\label{subsec:leakage-detection}
To detect leaking features $R_L$, we fit a relatively small model on corresponding rationales $\Phi_{\text{small}}: \mathcal{R} \rightarrow \mathcal{P}(\mathcal{Y})$ that predicts the label distribution from the rationale alone. We then calculate the attribution of its prediction to each input token using Integrated Gradient \citep{sundararajan2017axiomatic}. This token-level attribution is then averaged over the whole dataset to get global attribution for each token. \autoref{figure: pipeline} shows example attribution on different rationales. The highlight (in purple) denotes the token is critical to the model's prediction as its attribution is above a given threshold.

We would prefer the model to be \textit{less contextualized, less-trained, and smaller in size}. Our motivation is that smaller models won't be able to learn reasoning features, so their attributions tend to rely on shortcuts \citep{geirhos2020shortcut} like label leakage. In addition, attribution for smaller models is known to be more reliable and meaningful \citep{neely2022song}.

\subsection{Data Augmentation}
\label{subsec: data augmentation}

Given the original dataset $D = (X, R, Y)$, we generate partially counterfactual datasets
\begin{equation*}
    D^{e} = (X, R^{e}, Y), \ \forall e \in \mathcal{Y},
\end{equation*}
where editing is constrained to the spans identified in the previous step. Following MICE \citep{ross-etal-2021-explaining}, we train a label-associated sequence to sequence infiller. Specifically for the $i$-th datapoint $(x_i, r_i, y_i) \in D$, we create training data by processing the sentence $y_i \oplus x_i \oplus r_i$ with the same denoising objective as in T5 pretraining \citep{raffel2020exploring}, where a single sentinel mask replaces each consecutive span from gradient attribution (\S\ref{subsec:leakage-detection}) as inputs and the training objective is to predict masked spans.
Notice that we only mask within the rationale to control artificially introduced artifacts as discussed in the \autoref{Appendix: appendix-methods}. To augment a dataset $D$, we decode infilling for these masked spans for each $e \in \mathcal{Y}$ to create $D^{e}$. Notice that, unlike usual counterfactual generation, we do not necessarily require the generated rationale $r^{e}_i$ with prepending label $e$ to be highly associated with that label. In fact, the training of the editing model in MICE can be thought of as estimating

\begin{equation*}
    I_{\mathcal{V}}(Y \rightarrow R|X).
\end{equation*}
Therefore, only those parts of the rationale that are easy to predict from the label will be edited, and most other spans will likely remain the same (as shown in \autoref{figure: pipeline}), as the model predicts these spans mostly from the static $x_i$ instead of the flipped $y$. This is a desirable property in our case, as it leads to hyperparameters that are easy to tune, with which we apply a very aggressive removal threshold while still preserving most of the $R_R$.

% \subsection{Leakage Debiasing}
% \label{subsec: leakage debiasing}

\begin{table*}[t]
\small
\centering
\setlength\tabcolsep{3pt}
\begin{tabular}{p{0.11\textwidth}p{0.13\textwidth}p{0.22\textwidth}p{0.11\textwidth}p{0.22\textwidth}p{0.14\textwidth}}
\toprule
\textbf{Dataset}  & \textbf{Question}  & \textbf{\g{}}    & \textbf{\lky{}} & \textbf{\gl{}}  & \textbf{\va{}}                              \\ \midrule 
StrategyQA & \cellcolor{tabColor} Could Chuck Norris ride a horse? &  Chuck Norris is a person. Horses are bigger than people. People can ride horses. & \cellcolor{tabColor} The answer is True.  & Chuck Norris is a person. Horses are bigger than people. People can ride horses. The answer is True.&  \cellcolor{tabColor} Chuck Norris could ride a horse. \\  \bottomrule
\end{tabular}
\caption{Example of human-annotated and synthetic leaky rationales for fixed-label QA task. \lky{} verbatim leaks the label, \va{} is the declarative combination of the question and the given answer, and \gl{} is simply the concatenation of the \g{} and \lky{} rationales.} 
\label{table: rationale examples}
\end{table*}
\subsection{Computing \name{}}
\label{subsec: evaluation}

Given all the augmented datasets $\mathcal{D} \coloneqq \{D^e | \forall e \in \mathcal{Y}\}$, we train our evaluator on the joint of these datasets.\footnote{In reality, it is possible to reuse the original dataset $D$, as in a single batch the model sees corresponding data points from all datasets.} We learn an invariant predictor parameterized by $\Phi$ by applying the Invariant Risk Minimization (IRMv1) regularization \citep{arjovsky2020invariant}
\begin{equation}
\begin{aligned}
    \mathcal{L}(\Phi) =& \sum_{D^e \in \mathcal{D}} \mathcal{L}^e(\Phi) \\
    & \ \ + \lambda \|\nabla_{w|w=1.0}\mathcal{L}^e(w \cdot \Phi)\|^2,
\end{aligned}
\label{equation:irm}
\end{equation}
where $\mathcal{L}^{e}$ is the Empirical Risk Minimization (ERM) term on dataset $D^{e}$ and the second term is the IRM regularization to enforce $\Phi$ to be simultaneously optimal on all environments $D^e$. In our case, it is either the factual data or the counterfactual augmentation. $\lambda$ is a hyperparameter balancing the tradeoff between the predictive power (an ERM term) and the invariance of the predictor across different environments. When the underlying model is a classifier, \autoref{equation:irm} is directly applicable with normal cross-entropy loss. However, considering the increasing trend of directly decoding label sequence from pre-trained language models like T5 \citep{raffel2020exploring, nogueira2021investigating} and to better align with previous works \citep{chen-etal-2023-rev}, we use the following variant with label decoding:
\begin{equation}
\begin{aligned}
    \mathcal{L}(\Phi) &= \sum_{D^e \in \mathcal{D}} \mathbb{E}_{(x, r, y) \sim D^e}[-\log p_{\Phi}(y|x, r)] \\
    & \ + \lambda \|\nabla_{w|w=1.0}\log\frac{p_{\Phi}(y|x, r)}{\sum_{y' \in \mathcal{Y}} p_{\Phi}(y'|x, r)})\|^2.
    \label{equation: irm evaluation model}
\end{aligned}
\end{equation}
Notice that we still train to maximize probability over all possible strings in the ERM term to facilitate greedy decoding. In an ideal world with infinite data, this is equivalent to the conditional probability on the set of all possible labels. 

Additionally, we train a baseline model $\Theta$ by minimizing
\begin{equation}
    \mathcal{L}_{\text{Base}}(\Theta) = \mathbb{E}_{(x, r, y) \sim D} [-\log p_{\Theta}(y|x, \varnothing)].
    \label{equation: base evaluation model}
\end{equation}
% \daniel{"." between $x. \varnothing$ is a typo?}
As \autoref{eq:objective} can be decomposed into the difference between two conditional $\mathcal{V}$-entropy terms

\begin{equation*}
    S(R) = H_{\mathcal{V}} (Y | X) - H_{\mathcal{V}} (Y | R \setminus R_L \cup X).
\end{equation*}
With $\Theta$ and $\Phi$ from the same predictive family, the \name{} score $S(R)$ can thus be estimated as
\begin{equation}
    \begin{aligned}
    \hat{\mathcal{S}}(R)  =  \mathbb{E}_{(x, r, y) \sim D_{\text{test}}}\big[\log &p_{\Phi}(y|x, r) \\ -&\log p_{\Theta}(y|x, \varnothing) \big].
    \end{aligned}
    \label{equation: final score}
\end{equation}
Notice that the testing dataset $D_{\text{test}}$ has the same rationale distribution as the original dataset $D$.

If the model is simultaneously optimal for all $D^e \in \mathcal{D}$ \citep{arjovsky2020invariant}, it is ignorant \citep{xu2020theory} of the features that are different on $\forall D^{e} \in \mathcal{D}$ (otherwise, the predictive distribution will be different while the label distribution is the same). In our case, these features are supposed to be related to label leakages. We also discuss why naively masking out leaking parts does not work for free-text rationale evaluation in \autoref{Appendix: appendix-methods}.

% \yining{Do you think it is necessary to distinguish $D_{\text{train}}$ and $D_{\text{test}}$ for equation 10/11 and 12}
\begin{table*}[t]
\centering
\setlength{\tabcolsep}{5pt}
\begin{tabular}{lcccccccc}
\toprule
Rationales $\rightarrow$       &              & \multicolumn{3}{c}{Synthetic Leaky}                                                                                            & \multicolumn{4}{c}{Model Generated}                                                                                         \\ \cmidrule(l){3-5} \cmidrule(l){6-9} 
Metrics $\downarrow$                                     & \multicolumn{1}{l}{\g{}} & \multicolumn{1}{l}{\gl{}} & \multicolumn{1}{l}{\va{}} & \multicolumn{1}{l}{\lky{}} & \multicolumn{1}{l}{GPT-4} & \multicolumn{1}{l}{GPT-3.5} & \multicolumn{1}{l}{Llama2-7B} & \multicolumn{1}{l}{Flan-T5 Large} \\ \midrule
\name{}                     & 0.115                             & 0.119                     & 0.043                       & 0.038                            & 0.283                     & 0.253                       & 0.100                         & 0.061                             \\
\name{}$_{\textit{DeBERTa}}$  & 0.381                            & 0.348                     & 0.026                       & 0.024                             & 0.474                     & 0.459                       & 0.232                         & 0.132                             \\ \hdashline\noalign{\vskip 0.3ex}
\name{}$_{\textit{ablation}}$ & 0.121                            & 0.673                     & 0.505                       & 0.673                             & 0.501                     & 0.412                       & 0.155                         & 0.102                             \\ 
REV                                          & -0.038                           & 0.145                     & -0.005                      & 0.147                             & -0.066                    & -0.083                      & -0.035                        & -0.013                            \\
LAS                                          & 0.024                            & NaN                       & 0.131                       & NaN                               & 0.161                     & 0.215                       & 0.084                         & 0.027                             \\
RQ                                           & 0.138                            & 0.406                     & 0.376                       & 0.406                             & 0.316                     & 0.311                       & 0.121                         & 0.071                             \\ \hdashline\noalign{\vskip 0.3ex}
Human Eval & $\rule[0.5ex]{0.5em}{0.4pt}$ & $\rule[0.5ex]{0.5em}{0.4pt}$ & $\rule[0.5ex]{0.5em}{0.4pt}$ & $\rule[0.5ex]{0.5em}{0.4pt}$ & 2.69 & 2.32 & 1.16 & 0.39 \\
\bottomrule
\end{tabular}
\caption{Evaluation results on StrategyQA \citep{mor2021strategyqa} over synthetic and model-generated rationales. \textbf{Both \name{} and \name{}$_{\textit{DeBERTa}}$ rank \g{} and \gl{} high and close to each other, while they rank \va{} and \lky{} close to zero, indicating strong invariance to various label leakages. Additionally, \name{} and \name{}$_{\textit{DeBERTa}}$ consistently align the ranking of model-generated rationales with those of humans}.}
\label{table: evaluation results of strategyqa}
\end{table*}
\begin{figure*}
    \centering
    \includegraphics[width=\textwidth]{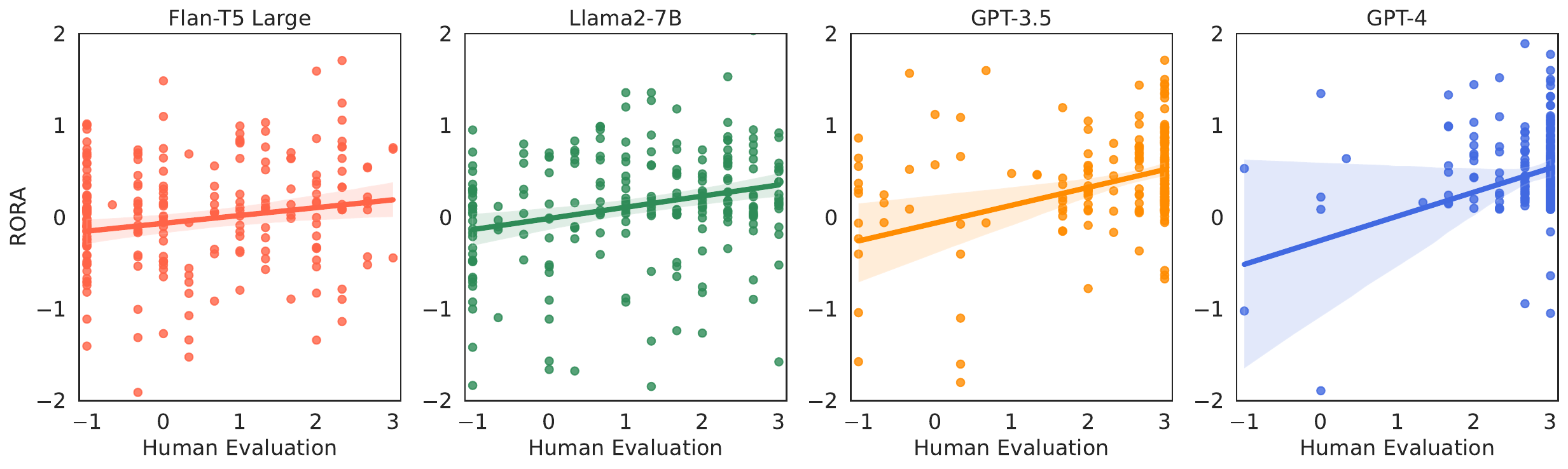}
    \caption{Linear regression on pointwise score correlation between human evaluation results and \name{} scores. Shades correspond to a 95\% confidence interval.}
    % \textbf{\name{} closely aligns with human evaluation, as the model size increases, data points tend to shift towards the upper right and become more densely clustered.}
    \label{figure: correlation plot}
\end{figure*}

\section{Experiments and Results}
% We outline our experimental setup by describing baseline metrics and model choices (\S\ref{subsec: baselines}), followed by the QA datasets and evaluation rationales (\S\ref{subsec: datasets and ratioanles}). Additional details of the experiment setup are provided in \autoref{Appendix: experiment setup}. 
\label{sec:expres}
To demonstrate the effectiveness of \name{}, we evaluate human-annotated, synthetic, and model-generated rationales on two distinct QA tasks. For synthetic data, we (1) write a sentence that verbatim leaks the label (\textbf{\lky{}}); (2) perturb the human-annotated rationales (\textbf{\g{}}) to include the \lky{} rationale (\textbf{\gl{}}); (3) create declarative rewrites (\textbf{\va{}}) of the question-answer pairs using a question rewriter \citep{chen-etal-2021-nli-models}.\footnote{\url{https://huggingface.co/domenicrosati/question_converter-3b}} \autoref{table: rationale examples} shows examples of these rationales.\footnote{For space concerns examples for ECQA are shown in \autoref{table: ecqa rationale examples} in Appendix~\ref{Appendix: experiment setup}} For all synthetic and human-annotated data, \textbf{a good metric for rationale quality should be able to evaluate \g{} and \gl{} reasonably close, while ranking them above \lky{} and \va{} rationales}. For model-generated rationales, \textbf{a good metric should evaluate rationales aligning with human judgments.} To create model-generated rationales of varying qualities, we choose GPT-4 \citep{openai2023gpt4}, GPT-3.5 \citep{ouyang2022training}, Llama2-7b \citep{touvron2023llama2} and Flan-T5 large \citep{chung2022scaling}, each being prompted with the same two demonstrations to generate a rationale for all question-answer pairs in StrategyQA (examples shown in Appendix~\ref{Appendix: model generated rationales}).

Without further specification, we use the gradient attribution \citep{sundararajan2017axiomatic} from FastText model \cite{joulin2017bag} to identify label-leaking tokens to mask. We use T5-base \citep{raffel2020exploring} to generate counterfactual data. Finally, two additional T5-base models are trained under \autoref{equation: irm evaluation model} and \autoref{equation: base evaluation model} as evaluation models to compute the \name{} score using \autoref{equation: final score}.\footnote{Note that the baseline evaluation model $\Theta$ does not condition on rationales, thus $\Theta$ can be trained once and the result of $\mathbb{E}_{(x,r,y)}\log p_{\Theta}(y|x,\phi)$ applies to the evaluation of various rationales types for the same dataset.}

We compare \name{} with three strong baselines: LAS \citep{hase-etal-2020-leakage}, RQ \citep{wiegreffe2022measuring}, and REV \citep{chen-etal-2023-rev}. All baselines use T5-base as the evaluation model. Note that the boundaries of these metrics differ depending on the choice of the $H_\mathcal{V}$ function in \autoref{equation: conditional v-information}. Specifically, LAS and RQ, which utilize accuracy, lie within the $[-1,1]$ interval. In contrast, \name{} and REV, employing the conditional $\mathcal{V}$-entropy (\autoref{equation: conditional v-entropy}), have a theoretical boundary of $[-\infty, \infty]$. Because of different scales and evaluation criteria, our analysis mainly focuses on the ranking over different rationale types, as well as the relative difference within each metric. We also do an ablation study with (\name{}$_\textit{ablation}$) with all the attributions, counterfactual generation and invariant learning removed. This is equivalent to the formulation in \autoref{equation: v-information} in which the evaluator $\Phi$ is directly trained on the given rationales and questions to maximize the likelihood of labels.

\begin{figure*}[t]
    \centering
    \includegraphics[width=\textwidth]{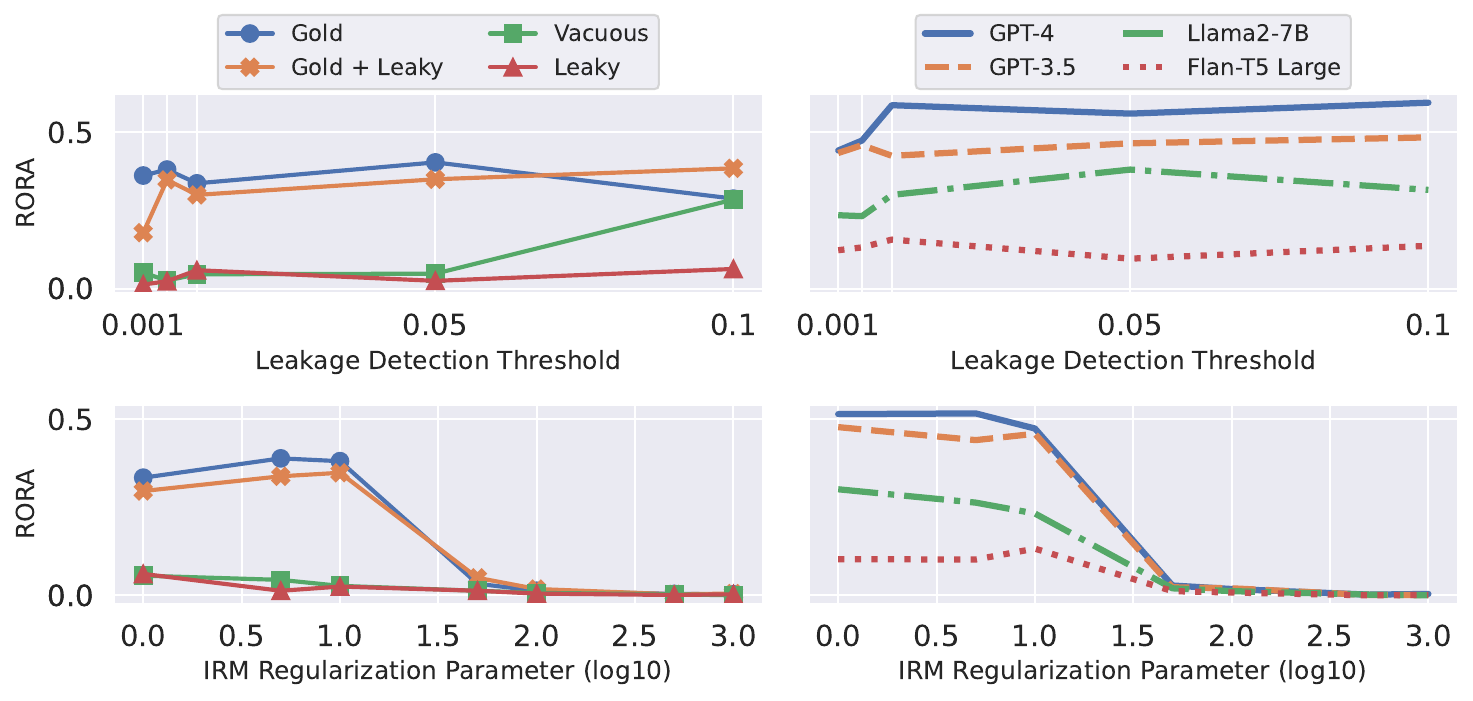}
    \caption{Sensitivity test results of \name{} on leakage detection threshold and IRM regularization parameter. \textbf{Decreasing threshold and increasing IRM regularization parameter help \name{} to better counteract label leakage. \name{} appears to be stable when the parameter was chosen from a relatively wide range}.}
    \label{figure: sensitivity analysis}
\end{figure*}
\subsection{Fixed-Label QA Task}
\label{subsec: results on fixed-label qa task}
We apply \name{} to StrategyQA \citep{mor2021strategyqa}, a multi-step commonsense reasoning dataset with fixed label set $\{\text{\texttt{True}}, \text{\texttt{False}}\}$. We concatenate the provided list of relevant facts to create \g{} rationales. Notice that \g{} rationale created this way has all the essential facts but does not explicitly define each reasoning step to be taken.

\autoref{table: evaluation results of strategyqa} shows evaluation results on human-annotated, synthetic, and model-generated rationales. Evidently, only full pipeline \name{} ranks \g{} and \gl{} the highest, much higher than scores of \va{} and \lky{}. This aligns with the human intuition that rationales with new, useful information should be ranked higher while restating labels should not affect scoring. However, all existing baselines violate the intuition and assign much higher scores to three synthetic leaky rationales than the \g{} rationale. We note that LAS is undefined for \lky{} and \gl{} rationales as it classifies all examples to leaked, thus causing a division-by-zero problem when averaging non-leaked examples, a problem discussed by \citet{pruthi2022evaluating}.

We observe similar trends with \name{}$_{\textit{DeBERTa}}$, which is a variation of \name{} with the final two T5-base evaluation models replaced by DeBERTaV3-large \citep{he2023debertav3} trained with \autoref{equation:irm} directly. This indicates that \name{} is model-agnostic, demonstrating comparable effectiveness when using different baseline models. On the other hand, \name$_{\textit{ablation}}$ is unable to provide comparable scores to \g{} and \gl{}, yet it assigns the highest to \lky{}, which again highlights the effectiveness of the \name{} approach in mitigating label leakage during evaluation.
\paragraph{Human Evaluation} We conduct a human evaluation for free-text rationales on Amazon Mechanical Turk. We annotate all 229 instances in our custom test split of StrategyQA. For each instance, we annotate rationales generated by all four models. For each rationale, the annotator is first asked to provide a binary judgment on whether the rationale supports the target label. If the answer is positive, the annotators are asked to judge how much new information the rationale provides beyond restating the question and the answer. This yields a natural 5-point scale to which we give scores in $\{-1, 0, 1, 2, 3\}$ respectively. Each rationale is annotated three-way redundant, and we average all human evaluation scores under the same model to obtain the final score for that particular rationale generation model. Additional human evaluation information is provided in Appendix~\ref{Appendix: human evaluation details}. The results are shown in \autoref{table: evaluation results of strategyqa}.

While RQ and \name{}$_\textit{ablation}$ also produce similar rankings as humans that give higher scores to larger LLMs, \name{} and \name{}$_\textit{DeBERTa}$ exhibit closer alignment with human evaluation and provide more fine-grained results. For instance, the \name{} score of GPT-3.5 is slightly lower than GPT-4 ($0.283 \rightarrow 0.253$, $0.474\rightarrow 0.459$), similar to human evaluation results ($2.69 \rightarrow 2.32$). Furthermore, pointwise $\mathcal{V}$-information estimated by \name{} for each rationale also has a decent correlation with human judgments, as shown in \autoref{figure: correlation plot} and \autoref{tab:correlation-with-human}.

\begin{table}[t]
\small
    \centering
    \begin{tabular}{lcc}
    \toprule
        \textbf{Method} & Pearson's $r$ & Spearman's $\rho$ \\
    \midrule
       REV \cite{chen-etal-2021-nli-models}  & 0.05 & 0.03 \\
       \name{} (Ours) & 0.36 & 0.37 \\
    \bottomrule
    \end{tabular}
    \caption{\name{} achieves a higher correlation with instance-wise human judgments.}
    \label{tab:correlation-with-human}
\end{table}

\subsection{Sensitivity Analysis}
\label{subsec: sensitivity analysis}

Considering that \name{} first detects leakage tokens based on a predetermined threshold and then leverages IRM to force the evaluation model to debiasing the label leakage, we study \name{}'s sensitivity towards the choice of the masking threshold as well as the IRM regularization parameter $\lambda$ in \autoref{equation: irm evaluation model}.

\paragraph{Masking Threshold} We test \name{}$_{\textit{DeBERTa}}$ across five masking thresholds on StrategyQA, where a lower threshold results in more aggressive detection. Results shown in \autoref{figure: sensitivity analysis} (first row) indicate that decreasing the threshold does induce stronger intervention, making the model ignorant of label leakage. For instance, the \name{} score on \va{} rationale drastically declines to 0 when the threshold decreases from 0.1 to 0.01. The \gl{} rationale is initially higher than \g{}, but after detecting more leakage tokens, it falls slightly below \g{} and keeps having the same trend. Additionally, \name{} consistently provides an expected ranking for model-generated rationales across different thresholds, demonstrating strong robustness for free-text rationale evaluation. As discussed in \ref{subsec: data augmentation}, we observe that \name{} is relatively robust to the threshold because most accidental over-masking is likely to be filled with identical tokens, provided that the token does not have a strong correlation with the label.

\paragraph{IRM Coefficient $\lambda$} The IRM regularization coefficient controls the strength of the invariant learning. A higher value forces the evaluation model to focus more on the shared features between the leaked and non-leaked environments while potentially sacrificing performance to the single environment. We test \name{}$_{\textit{DeBERTa}}$ over seven different regularization parameters ranging from 1 to 1000 on StrategyQA and results are shown in \autoref{figure: sensitivity analysis} (second row). The increase of parameters from 1 to 20 leads to the convergence of lines \g{} and \gl{}, while still being higher than \va{} and \lky{}. This demonstrates that invariant learning is helpful for evaluation models to mitigate label leakage in rationales. However, exceedingly large regularization parameters, such as those higher than 100, can make evaluation models unlearn features in leaked and non-leaked environments, resulting in declined scores close to 0. We also observe that \name{} provides a correct ranking for model-generated rationales over a wide and reasonable range of regularization parameters.

\subsection{Open-Label QA Task}
\label{subsec: results on open-label qa task}

We further test \name{} on CommonsenseQA \citep{talmor-etal-2019-commonsenseqa}, where we evaluate both ECQA \citep{aggarwal-etal-2021-explanations} and COS-E v1.11 \citep{rajani2019explain}. CommonsenseQA adopts a multiple-choice setting where the label set for each problem differs from one another. The model usually needs contextualization to make meaningful predictions when the inputs are linearized into plain text. To have a minimally contextualized attribution model, we use LSTM \citep{hochreiter1997long} bi-encoder model to separately encode the question and choices and select the final label by calculating the dot product. We perturb ECQA rationales to generate \gl{}, and other leaky rationales are created similarly as in the previous task.

\autoref{table: evaluation results of ecqa} shows that \name{} again exhibits strong robustness to label leakages. It ranks \g{} and \gl{} rationales the highest and gives \va{} and \lky{} fairly low scores, while existing baselines are hard to achieve. Additionally, \name{} also finds that the quality of CoS-E is inferior to that of ECQA, consistent with the evaluation results from other baselines \citep{chen-etal-2023-rev} and human observations \citep{aggarwal-etal-2021-explanations}. While REV \citep{chen-etal-2023-rev} also ranks all synthetic rationales similarly to \name{}, it is because it only takes the rationale as input, and on an open-label dataset, this singlehandedly does not deterministically leak the label. However, our further experiments reveal other potential problems with REV (\autoref{Appendix: further discussion on rev and label leakage}). On the contrary, \name{} is not susceptible to such adversarial, as \name{} evaluator already has access to everything in the question inputs.
% (0.436 according to our experiment)

\begin{table}[t]
\small
\centering
\setlength\tabcolsep{4pt}
\begin{tabular}{@{}lccccc@{}}
\toprule
Rationales                                       & \multicolumn{2}{c}{Human-Annotated}          & \multicolumn{3}{c}{Synthetic Leaky} \\ \cmidrule(l){2-3} \cmidrule(l){4-6} 
Metrics $\downarrow$                                           & CoS-E & \g{} & \gl{}   & \va{}  & \lky{}  \\ \midrule
\name{}                          &   0.398   & 1.30    &       1.33       &     0.325    &     0.025    \\
REV                                               &   0.128    &       0.284       &    0.271     &   0.013    &        -0.010        \\
LAS                                               & 0.198 & 0.365        & 0.373   & 0.425    & 0.506          \\
RQ                                                & 0.235 & 0.425        & 0.432   & 0.402    & 0.432          \\ \bottomrule
\end{tabular}
\caption{Evaluation results on ECQA \citep{aggarwal-etal-2021-explanations} over different rationales, including human-annotated rationales from CoS-E \citep{rajani2019explain}. \textbf{\name{} shows strong robustness in evaluating leaky rationales for open-label QA questions}.}
\label{table: evaluation results of ecqa}
\end{table}

\section{Related Work}
\label{sec: related works}
\subsection{Gradient-based Attribution}
\label{subsec:gb-attribution-methods}
Gradient-based attribution methods have been very popular in interpreting neural predictions. Comparative to perturbation \citep{zeiler2014visualizing, zintgraf2017visualizing} or Shapley-value-based attributions \citep{shapley1953value, lundberg2017unified, strumbelj2010efficient, chen2020generating}, variants of gradient-based attribution \citep{sundararajan2017axiomatic, shrikumar2017learning} can be very efficient to compute as the number of samples needed to generate attribution does not depend on the number of features \citep{ancona2018towards, lyu2024towards}. Therefore, in this work, we use integrated gradient (IG) \citep{sundararajan2017axiomatic} attribution for the small and small model for label leakage identification because IG is a generally applicable method that has been shown to provide good explanations for NLP tasks \citep{pruthi2022evaluating}, while more easily accessible methods are either non-applicable due to non-linearity being employed in the computation process \citep{ancona2018towards}, or requiring specific model structures while still generating questionable attributions \citep{jain2019attention, wiegreffe-pinter-2019-attention, pruthi2019learning, bibal2022attention, ethayarajh2021attention, liu2022rethinking}.

\subsection{Contrastive Generation}
\label{subsec: contrastive generation}
% \citet{ross-etal-2021-explaining} introduce MICE, a method to get fluent contrastive explanations by generating edits to the input explanation that change the predictor's output from the original prediction to a contrasting one. CREST \citep{treviso-etal-2023-crest}, on the other hand, first masking relevant tokens and then infilling spans conditioned on the contradicting label. \name{} adapts the strategy of MICE to generate contrastive leakage rationales.

To better understand model decision boundary, \citet{gardner2020evaluating} propose to evaluate models with contrast sets, which consist of test instances perturbed in small but meaningful ways that change the gold label. As human supervision in terms of contrast sets helps evaluate and improve classifiers \citep{kaushik2019learning, chen2023models}, a couple of methods have been proposed to automatically generate counterfactuals as a form of explanation \citep{li2020linguistically, yang-etal-2020-generating, wu-etal-2021-polyjuice, dixit2022core, ross-etal-2022-tailor, jacovi-etal-2021-contrastive}. In particular, SMG \citep{sha-etal-2021-controlling}, MICE \citep{ross-etal-2021-explaining}, and CREST \citep{treviso-etal-2023-crest} adopt a select-and-edit two-step process to generate label-changing counterfactuals. In this work, we adapt MICE to generate partial counterfactuals, where only label-leaking parts of a rationale that spuriously correlate to the label get edited.

\subsection{Invariant Learning}
\label{subsec: invariant learning}
Invariant Risk Minimization (IRM) \cite{arjovsky2020invariant, ahuja2020invariant} is a recently proposed framework for learning invariant predictors to spurious correlations. While most of these works primarily focused on theoretical findings, simple models, and toy datasets, in a notable departure \citet{NEURIPS2021_972cda1e} first tests IRM on natural language inference, pointing out that IRM has a potential edge over empirical risk minimization on tasks or datasets with stronger and more prevalent biases and larger data size. This work combines IRM with counterfactual data augmentation to train an evaluator agnostic of label leaking features.

\section{Conclusions}

We introduce \name{}, an innovative approach designed to enhance the accuracy and robustness of rationale quality evaluation amidst the challenges posed by label leakage. 
Our method stands out for its adaptability across diverse model architectures and hyperparameter configurations, delivering consistent rationale quality evaluations that align more closely with human evaluation.

\section*{Limitations}
While \name{} addresses the problem of label leakage, we focus on the evaluation of the informativeness of a rationale. While an easy-to-follow reasoning path in the rationale seems to boost evaluation score, \name{} is not sensitive to other aspects of rationale quality. For example, a good rationale should also be factual \citep{prasad2023receval} and consistent \citep{chen2023two}, which is not directly addressed in this work.

\section*{Acknowledgement}
A.L. was partially funded by a JHU Discovery Award and a JHU Institute for Assured Autonomy (IAA) seed grant. 
D.K. is supported in part by ONR grant N00014-241-2089, and generous gifts from Amazon and the Allen
Institute for AI. This work has also been partially supported by the U.S. National Science Foundation under grant 2204926. Any opinions, findings, and conclusions or recommendations expressed in this article are those of the authors and do not necessarily reflect the views of the National Science Foundation.
Finally, we are grateful to the anonymous reviewers for constructive feedback for improving this work. 

% Entries for the entire Anthology, followed by custom entries
\bibliography{custom}
\bibliographystyle{acl_natbib}

\onecolumn

\clearpage
\appendix
\begin{center}
{\LARGE \textbf{Supplemental Material}}
\end{center}

\begin{center}
\begin{tabular}{@{}ll@{}}
\toprule
Appendix & Contents \\ \midrule
\autoref{Appendix: appendix-methods}       & Justifying Partial Counterfactual Augmentation        \\
\autoref{Appendix: further discussion on rev and label leakage} & Further Discussion on REV and Vacuous Rationale \\
\autoref{Appendix: experiment setup}  & Additional Details of Experimental Setup  \\
% \autoref{Appendix: experiment results} & Additional Details of Experiment Results \\ \bottomrule
\bottomrule
\end{tabular}
\end{center}

\section{Justifying Partial Counterfactual Augmentation}
\label{Appendix: appendix-methods}

This section provides intuitive explanations of why other simple alternatives may not work. We consider the following alternatives:

\begin{itemize}
    \item Mask out all the tokens attributed as leakage by the smaller model.
    \item IRM training with the environment $D$ and $D^{\text{Mask}}$, which we denote by $e_1$ and $e_2$ respectively.
\end{itemize}

\subsection{Training on Masked Inputs}

The major problem is that masking may introduce new artifacts. For instance, in English, negation is applied to the auxiliary verb. Therefore, if the model identifies a discrepancy in tense and a masked token preceding the main verb, it's highly probable that the masked token represents negation. In our pilot experiments, we found that models are capable of utilizing these traits. Also, attribution calculated from positive and negative labels might be different, a pattern the model might learn to exploit. That's why in \name{}, the attribution is calculated with the global weighting. Partially counterfactual generation makes the sentence more natural, which helps reduce artifacts introduced.

\subsection{Using Masked Inputs as an Environment}

We hereby show that invariant learning with a masked environment does not force the model to unlearn $R_{L}$, as described in \autoref{eq:objective}. In general, creating scenarios where one environment includes a non-essential characteristic and the other does not will not result in an Invariant Risk Minimization (IRM) model that achieves comparable loss levels regardless of whether these non-essential characteristics exist. To demonstrate this, we create a minimal example from the causal graph from LAS \citep{hase-etal-2020-leakage}, where each simulation variable is replaced with its actual variable as shown in \autoref{fig:las-causal}.

\begin{figure}[htbp]
    \centering
    \begin{tikzpicture}[
    roundnode/.style={circle, draw=black!60, fill=white!5, very thick, minimum size=7mm},
    observedroundnode/.style={circle, draw=black!60, fill=black!20, very thick, minimum size=7mm}
    ]
    %Nodes

    \node[observedroundnode] (rationale)                              {$R$};
    \node[roundnode]         (leaky)       [above right= 0.2em and 1em of rationale]       {$R_{L}$};
    \node[roundnode]         (reasoning)   [below=1em of leaky]           {$R_{R}$};
    \node[observedroundnode] (label)       [below right=0.2em and 1em of leaky]           {$Y$};
    \node[observedroundnode] (other)       [right=1em of label]           {$X$};
    \node[roundnode]         (environment) [above= 1em of leaky]           {$E$};

    %Lines
    \draw[-Triangle] (rationale.north) to[bend left] (leaky.west);
    \draw[-Triangle] (rationale.south) to[bend right] (reasoning.west);
    \draw[-Triangle] (reasoning.east) to[bend right] (label.south);
    \draw[-Triangle] (leaky.east) to[bend left] (label.north);
    \draw[-Triangle] (other.west) -- (label.east);
    \draw[-Triangle] (environment.south) -- (leaky.north);
    
    \end{tikzpicture}

    \caption{A causal graph showcasing the generative story of the label $Y$ of an instance, where correctness indicator variables from LAS \cite{hase-etal-2020-leakage} are replaced by an actual variable.}
    \label{fig:las-causal}
\end{figure}

Each variable corresponds to its designation in \S\ref{subsec: evaluating label-leaking rationales}. For simplicity, let us consider the following semantics of each random variable: $R$ is a free-text rationale taking values from all possible rationale documents; $R_{L}$ is a ternary variable taking values from $r_l \in \{\text{True}, \text{False}, \text{UNK}\}$, where UNK denotes masks; $R_{R}$ is a binary indicator of $r_{r} \in \{0, 1\}$; and $X$ is the binary indicator of $x \in \{0, 1\}$ indicating the binary choice innate to the original question. We assume the following conditional probability:

\begin{align*}
    p(Y = \text{True}\ |\ R_{L} = \text{True}, X, R_{R}) &= 1,\\
    p(Y = \text{True}\ |\ R_{L} = \text{False}, X, R_{R}) &= 0,\\
    p(Y = \text{True}\ |\ R_{L} = \text{UNK}, X = R_{R}) &= 0.9,\\
    p(Y = \text{True}\ |\ R_{L} = \text{UNK}, X \neq R_{R}) &= 0.
\end{align*}
In this case, the relationship $R_{L} \Rightarrow Y$ \textit{is causal}, in the sense that the model can discern which environment the prediction is made in by looking at $R_{L}$. Specifically, assuming that we can identify the leakage part correctly with the application of IG \citep{sundararajan2017axiomatic}, then the environments will be

\begin{align*}
    e_1 &\coloneqq \Big\{(R^{e_1}, X^{e_1}, Y^{e_1})\ |\ p(R_{L}^{e_1} = \text{UNK}|R^{e_1}) = 1\Big\},\\
    e_2 &\coloneqq \Big\{(R^{e_2}, X^{e_2}, Y^{e_2})\ |\ p(R_{L}^{e_2} = \text{UNK}|R^{e_2}) = 1\Big\}.
\end{align*}
Now suppose we have infinite computation, and it is possible to learn any representation $\Phi: \mathcal{R} \times \mathcal{X} \rightarrow \mathbb{R}^d$, and a linear classification head $\tilde{w}: \mathbb{R}^d \rightarrow \mathcal{P}(\mathcal{Y})$ (which outputs a single value between [0, 1]). We are considering the risk function $F$ that is common in classification settings:

\begin{align*}
    F(\tilde{w}, \Phi) = 
        \mathbb{E} 
        \Big[ - \mathbb{I}[y = \text{True}] \cdot \log 
            \big(
                \tilde{w}\circ \Phi (x)
            \big) - \mathbb{I}[y = \text{False}] \cdot \log \big( 1 - \tilde{w}\circ \Phi (x)
            \big)
        \Big].
\end{align*}
Suppose we have a representation that fully recovers all the causal components

\begin{align*}
    \Phi(r, x) = \Big\langle
        \mathbb{I}[x = q] \cdot \mathbb{I}[r_l = \text{UNK}],
        \mathbb{I}[r_l = \text{True}]
    \Big\rangle,
\end{align*}
Then it is easy to see that $\tilde{w}(z) = v^tz$ where $v = \langle0.9, 1\rangle$ will be the optimal classifier across both environments $e \in \{e_1, e_2\}$. It can be verified that

\begin{align*}
    \lim_{v' \rightarrow v} \nabla_{w|w=1.0} F(\tilde{w}, \Phi) = 
    -p(Y = \text{True})\frac{1}{v^T\Phi(x)}+ p(Y = \text{False}) \frac{1}{1 - v^T\Phi(x)}.
\end{align*}
Which, given our $\tilde{w}$ and $\Phi$, does not penalize the predictions. But does this invariant predictor satisfy our desirable property that it will rank instances disregarding whether there is a label leakage? The answer is no, as here, the instances with label leakage will get zero losses, while those without will get non-zero losses. This highlights the importance of generating a counterfactual environment with natural-looking data with a different label association $p(Y|R_{L})$, like what we did for \name{}.

\section{Further Discussion on REV and Vacuous Rationale}
\label{Appendix: further discussion on rev and label leakage}
REV \citep{chen-etal-2023-rev} evaluates a rationale based on a vacuous rationale $B$. We argue that it does not follow the natural way of how humans evaluate a rationale, where we typically need to read and understand the question before the evaluation. Therefore, an adjusted version of REV would be 
$$\text{REV}'(R) = I_{\mathcal{V}}(X,R \rightarrow Y | B).
$$ 
However, the compositionality nature of language \citep{pelletier1994principle} implies that in many cases 
\begin{equation}
\begin{aligned}
    I_\mathcal{V}(X, R &\rightarrow Y|B) - I_\mathcal{V}(R \rightarrow Y|B) \neq \\
    &I_\mathcal{V}(X, R \rightarrow Y| \varnothing) - I_\mathcal{V}(R \rightarrow Y| \varnothing).
    \label{eq:counter-x}
\end{aligned}
\end{equation}
This reveals another potential weakness of REV, which makes restating the question an
incredibly high-scoring rationale (0.436 according to our experiment). Indeed, this is counter-intuitive as human evaluators should always have access to $X$ but not $B$. On the contrary, \name{} does not have this issue as it already conditions on $X$.

\section{Experiment Setup}
\label{Appendix: experiment setup}

\subsection{Models and Datasets}
We access all models through Huggingface Transformers \citep{wolf2019huggingface}. We split the StrategyQA \citep{mor2021strategyqa} train set into our custom train, validation, and test sets and then create model-generated rationales for each set. For ECQA \citep{aggarwal-etal-2021-explanations} we use their original splits.

% \begin{table}[t]
% \centering
% \begin{tabular}{@{}lccc@{}}
% \toprule
% Dataset         & \# Train & \# Validation & \# Test \\ \midrule
% StrategyQA      & 1374     & 458           & 458     \\
% Model-Generated & 1374     & 458           & 458     \\
% ECQA            & 7598     & 1090          & 2194    \\
% CoS-E           & 9741     & 1221          &         \\ \bottomrule
% \end{tabular}
% \caption{Summary statistics of the datasets.}
% \end{table}

\subsection{Training Hyperparameters}
We train each small model for gradient attribution calculation with a learning rate of $1e-1$ and batch size of 256. For leakage token detection, we choose threshold $0.005$ for \name{}$_{\textit{DeBERTa}}$ and $0.01$ for \name{}, or use top-1 strategy. To determine the appropriate threshold, we assess the output on a small sample of training data, ensuring that at least the majority of the samples receive more than one masking beyond that from the top-k selection. We train each counterfactual data generator using a learning rate of $1e-4$ with batch sizes 8. For invariant learning, we use IRM regularization $\lambda = 10$ for both \name{} and \name{}$_{\textit{DeBERTa}}$ and train each evaluation model with a learning rate of $1e-4$ and batch size of 64. For each training, we use optimizer AdamW \citep{loshchilov2019decoupled} from Transformers \citep{wolf2019huggingface} and set maximum epochs as 20 with early stopping.

\subsection{Human Evaluation Details}
\label{Appendix: human evaluation details}

\begin{figure}[htbp]
    \includegraphics[width=\textwidth]{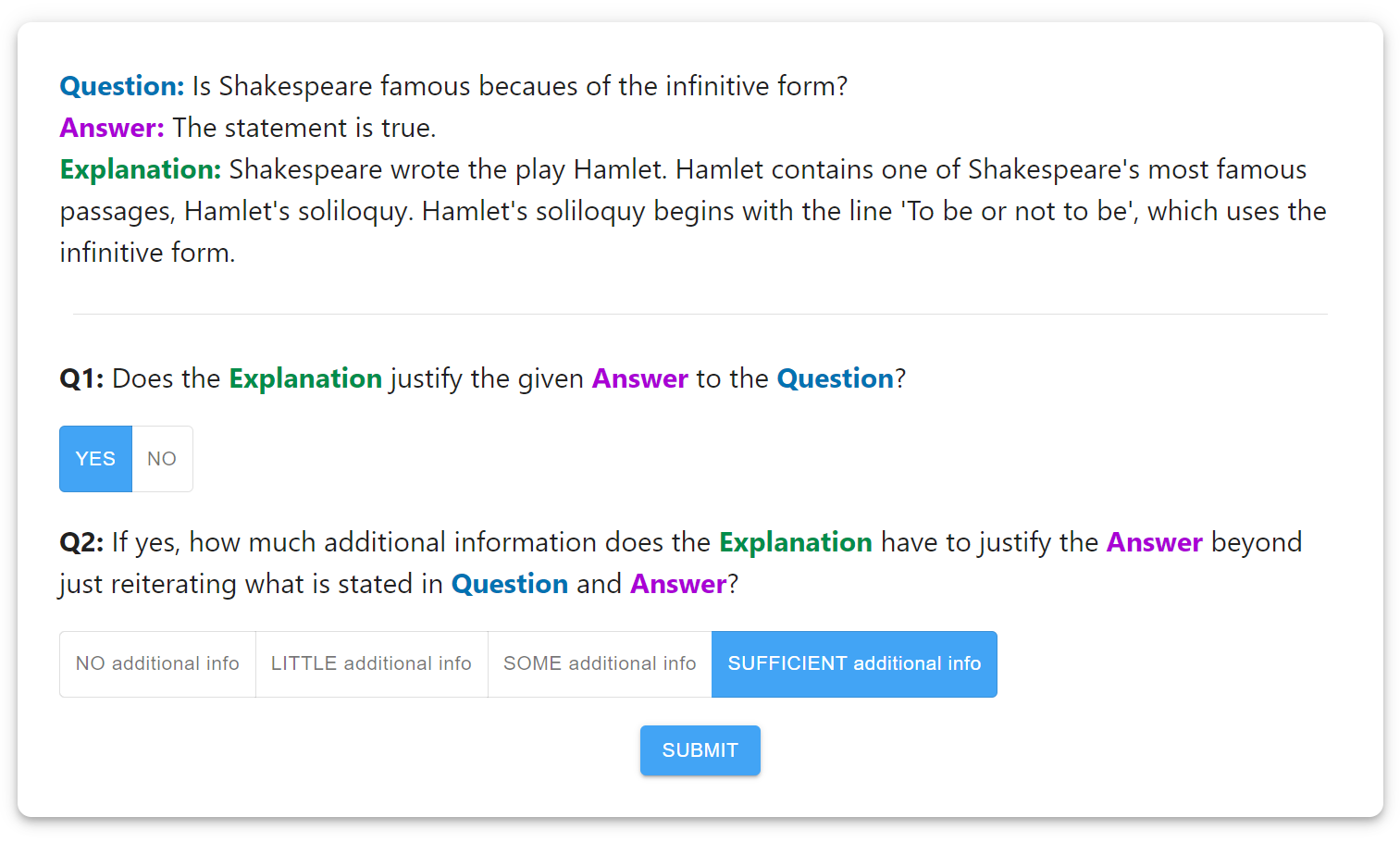}
    \caption{The interface to our human validation hit.}
    \label{fig:question-template}
\end{figure}

\begin{figure}[htbp]
    \includegraphics[width=\textwidth]{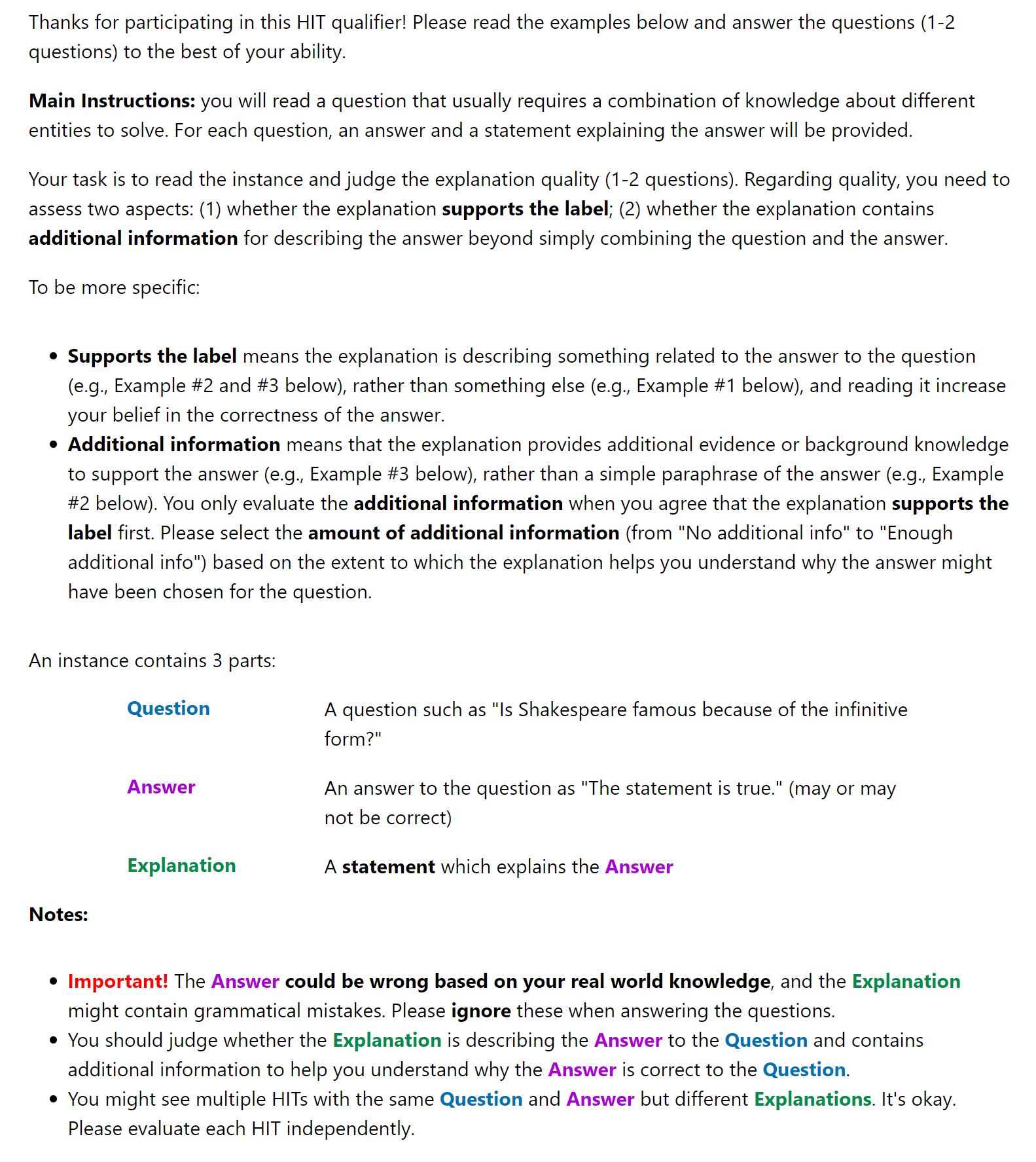}
    \caption{The detailed instruction we give the annotators. Notice that we specifically instructed them to not judge the factuality based one their real-world knowledge.}
    \label{fig:instructions}
\end{figure}

We conduct a human evaluation of model-generated rationale on Amazon Mechanical Turk\footnote{\url{https://www.mturk.com/}}. The quality of each rationale is judged by three individual annotators, who must pass a qualification test described in \cite{chen2020uncertain}. Annotators have been paid \$0.05 per rationale, which amounts to an hourly wage of about \$10. \autoref{fig:question-template} shows an example of our annotation interface. For each rationale, We present annotators with a question, an answer, and the model-generated rationale (explanation) and ask them the following questions:
\begin{enumerate}
    \item \textit{Does the Explanation justify the given Answer to the Question?} We offer two options for them to choose from: ``Yes'' and ``No''. 
    \item If they answer ``Yes'', we further ask them \textit{how much additional information does the Explanation have to justify the Answer beyond just reiterating what is stated in Question and Answer?} and to pick one option from ``No Additional Info'', ``Little Additional Info'', ``Some Additional Info'' and ``Sufficient Additional Info'' which are corresponding to a 4-point Likert-scale (0/1/2/3). If the annotator answers ``No'' for the first question, we label the rationale a score of -1.
\end{enumerate}   
Then, we average all three annotators' scores to get the human evaluation score for that single rationale. For detailed instructions, readers can refer to \autoref{fig:instructions}.

\subsection{Model Generated Rationales}
\label{Appendix: model generated rationales}
\autoref{table: example of model-generated rationales} shows examples of rationale generated by different models.

\newpage

\begin{table*}[htp]
\small
\centering
\setlength\tabcolsep{3pt}
\begin{tabular}{p{0.08\textwidth}p{0.12\textwidth}p{0.25\textwidth}p{0.10\textwidth}p{0.25\textwidth}p{0.11\textwidth}}
\toprule
\textbf{Dataset}  & \textbf{Question}  & \textbf{\g{}}    & \textbf{\lky{}} & \textbf{\gl{}}  & \textbf{\va{}}                              \\ \midrule 
ECQA & \cellcolor{tabColor} What town is known for disco? Options: op\_1 rio de janeiro op\_2 new york op\_3 town op\_4 germany op\_5 hotel & Rio de Janeiro is a huge seaside city in brazil, famed for its Copacabana and Ipanema beaches. The town Rio de Janeiro is known for disco. New York city is famous for Central park, Statue of liberty, Empire state building and other such places and not for disco. Town is larger than a village and generally smaller than a city. Town is not the name of a place where disco is famous for. Germany is the country and not a town known for disco. Hotel is not the town known for disco. & \cellcolor{tabColor} The answer is Rio de Janeiro. & Rio de Janeiro is a huge seaside city in brazil, famed for its Copacabana and Ipanema beaches. The town Rio de Janeiro is known for disco. New York city is famous for Central park, Statue of liberty, Empire state building and other such places and not for disco. Town is larger than a village and generally smaller than a city. Town is not the name of a place where disco is famous for. Germany is the country and not a town known for disco. Hotel is not the town known for disco. The answer is Rio de Janeiro. & \cellcolor{tabColor} Rio de Janeiro is known for disco.\\ \bottomrule
\end{tabular}
\caption{Example of human-annotated and synthetic leaky rationales for open-label QA task. \lky{} verbatim leaks the label, \va{} is the declarative combination of the question and the given answer, and \gl{} is simply the concatenation of the \g{} and \lky{} rationales.} 
\label{table: ecqa rationale examples}
\end{table*}

\begin{table}[htbp]
\centering
\begin{tabular}{p{0.1\textwidth}p{0.2\textwidth}p{0.5\textwidth}}
\toprule
\textbf{Model} & \textbf{Question and Label}                                                                & \textbf{Generated Rationale}                                                                                                                                                                                                                 \\ \midrule
GPT-4          & Question: Could the Great Wall of China connect the Dodgers to the White Sox? Answer: True & The Great Wall of China is 1,948 km long. The Yankees are based in New York City. The Dodgers are in Los Angeles and play in Los Angeles. The White Sox are based in Chicago and play in Chicago. The answer: yes.                           \\
& & \\
GPT-3.5        & Question: Could the Great Wall of China connect the Dodgers to the White Sox? Answer: True & No rationale can be provided to explain this answer because it is not accurate. The Great Wall of China is located in China and does not connect the Dodgers (a baseball team in Los Angeles) to the White Sox (a baseball team in Chicago). \\
& & \\
Llama2-7B      & Question: Could the Great Wall of China connect the Dodgers to the White Sox? Answer: True & The Great Wall of China is a long wall. The Dodgers and White Sox are baseball teams.                                                                                       \\
& & \\
Flan-T5 Large  & Question: Could the Great Wall of China connect the Dodgers to the White Sox? Answer: True & The Great Wall of China is 1,948 km long. The Yankees are based in New York City. The Dodgers are in Los Angeles and play in Los Angeles. The White Sox are based in Chicago and play in Chicago. The answer: yes.                           \\ \bottomrule
\end{tabular}
\caption{Examples of model-generated rationales}
\label{table: example of model-generated rationales}
\end{table}

\end{document}